\documentclass[runningheads]{llncs}
\usepackage{graphicx}
\usepackage{setspace}
\usepackage{url}

\begin{document}
\title{RECKONition: a NLP-based system for Industrial Accidents at Work Prevention\thanks{This work has been funded by INAIL within the BRiC/2018, ID09 framework, project RECKON. Many thanks to Prof. Enrico Cagno from the Department of Management, Economics and Industrial Engineering (DIG) and Prof. Francesco Braghin from the  Department of Mechanical Engineering for their contribution.}\\~\\
\large DISCUSSION PAPER}
\titlerunning{RECKONition: a NLP-based system}
%
\author{Patrizia Agnello$^a$, Silvia M. Ansaldi$^a$, \\ Emilia Lenzi$^b$, Alessio Mongelluzzo$^b$, Manuel Roveri$^b$}
\authorrunning{P. Agnello et al.}
%
\institute{$^a$Inail - Dipartimento Innovazioni Tecnologiche, Italy\\
$^b$Politecnico di Milano, Italy}

\maketitle              
\begin{abstract}
Extracting patterns and useful information from Natural Language datasets is a challenging task, especially when dealing with data written in a language different from English, like Italian. Machine and Deep Learning, together with Natural Language Processing (NLP) techniques have widely spread and improved lately, providing a plethora of useful methods to address both Supervised and Unsupervised problems on textual information. We propose RECKONition, a NLP-based system for Industrial Accidents at Work Prevention. RECKONition, which is meant to provide Natural Language Understanding, Clustering and Inference, is the result of a joint partnership with the Italian National Institute for Insurance against Accidents at Work (INAIL). The obtained results showed the ability to process textual data written in Italian describing industrial accidents dynamics and consequences.

\end{abstract}
\section{Introduction}
\label{sct:introduction}
Despite the ever-growing awareness and the advances in the technology and the procedure, the problem of accidents at work represents a relevant issue from both the social and economic point of view. In more detail, accidents at work refer to damages to the health of the worker induced by accidents correlated with the working activities. Unfortunately, such accidents can result into serious damages, such as temporary or permanent reduction or loss of working capacity, or even death. In Italy, this issue is particularly perceived as very relevant since in last few years the number of accidents at work is larger than 500.000 per year, among which more that 1000 resulted in the death of the worker. Addressing this issue in Italy is the primary goal of INAIL, which is National Insurance Institute for Industrial Accidents and Occupational Diseases. More specifically, INAIL’s objectives are the reduction of injuries, the protection of workers performing hazardous jobs, and the 
facilitation of the return to work of people injured at workplace. To achieve these relevant and challenging objectives, INAIL explores novel technological and societal solutions and paradigms to provide an integrated system of protection, ranging from preventive actions at the workplace to  medical services and financial assistance.

Many Natural Language Processing (NLP) applications and solutions can be found in the literature to tackle tasks like Sentiment Analysis or Text Classification, for instance~\cite{ref_sentiment}. However, when it comes to the Italian language, and injury prevention in particular, there are no suited solutions available. In fact, some of the libraries that support the implementation of the most novel NLP techniques also allow users and researchers to upload and share the models (e.g., BERT~\cite{ref_bert}) they build and train for their specific tasks\footnote{We refer to those models hubs like HuggingFace (\url{https://huggingface.co/models}) or TensorFlow Hub (\url{https://tfhub.dev/})}. Here we can find some models which were fine tuned on Italian datasets too, still not being suited to the case of injuries and industrial accidents.
In this path, the aim of this paper is to introduce RECKONition, a novel Natural Language Processing-based system to extract knowledge from textual descriptions of accidents at work aiming at defining industrial accidents preventive actions at the workplace. The RECKONition system comprises three different modules, i.e., Association Rule Generation, Textual Description Clustering and Textual Description Inference, able to extract relationships between accident events and create groups of homogeneous accident descriptions representing the input for the definition of preventive actions. The RECKONition system has been successfully applied to a real-world database storing the textual descriptions of accidents occurring in Northern Italy from 2013-2018. 

The paper is organized as follows. In Section~\ref{sct:reckonitionsystem} we introduce RECKONitions system and its three algorithmic modules. In Section~\ref{sct:associationrulesmining},~\ref{sct:naturallanguageclustering}, and~\ref{sct:naturallanguageprediction} respectively, we detail our implementation of the Association Rules Mining, Natural Language Clustering, and Natural Language Prediction modules. In Section~\ref{sct:discussion_conclusion} we summarize our work and discuss on the possible next steps and improvements
%
%
%
%
\vspace{-0.5cm}
\section{The RECKONition system: extracting knowledge from textual descriptions of accidents at work}
\label{sct:reckonitionsystem}
\vspace{-0.4cm}
The architecture of the RECKONition system, which is described in Fig. \ref{fig:arch}, comprises three different modules: Association Rule Generator, Textual Description Clustering and Textual Description Inference. All these models receive in input a set of textual descriptions of accidents at work and provide in output \textit{association rules} to discover relationships between terms, \textit{clusters of textual descriptions} highlighting groups of accidents at work sharing similarities in their descriptions, and  \textit{next sentence predictions} from the textual description of the accidents. All the outputs of the RECKONition system represent valuable tools for the INAIL expert to gain knowledge about occurred accidents at work and define prevention policies. The three modules of RECKONition system are detailed in the next sections.
\begin{center} 
    \includegraphics[width=0.8\textwidth]{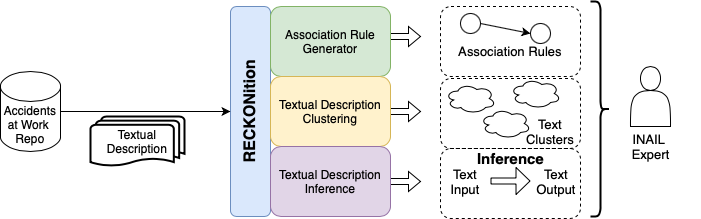}
    \vspace{-1cm}
    \begin{figure}[h]
    \caption{The Architecture of the RECKONition system.}
    \label{fig:arch}
    \end{figure}
\end{center}
\vspace{-1cm}
\section{Association Rules Mining}
\label{sct:associationrulesmining}
Association Rules (ARs) allow to discover relations between variables in large datasets and to graphically show such dependencies in a convenient representation which can be easily interpreted by humans~\cite{ref_association_rules} (we show in Figure~\ref{fig:arsample} an example of an AR representation).
\begin{center}
    \includegraphics[width=0.5\textwidth]{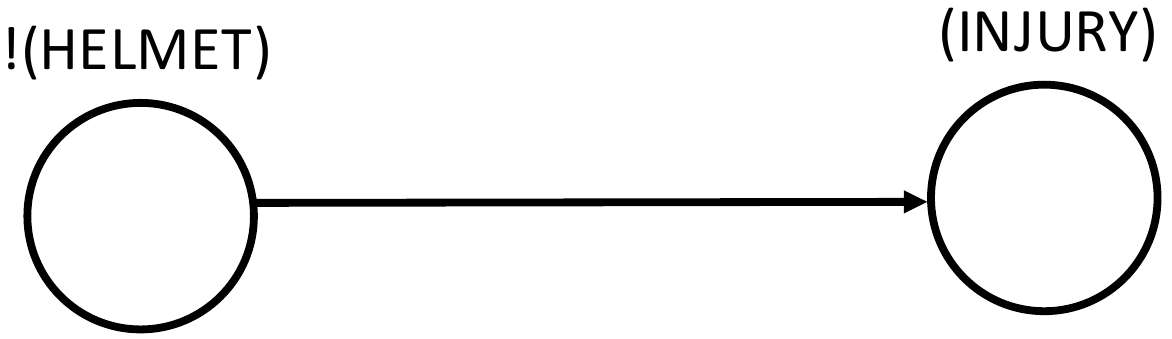}
    \vspace{-1cm}
    \begin{figure}[h]
    \caption{An example of a negative association rule graphical representation.}
    \label{fig:arsample}
    \end{figure}
\end{center}
\vspace{-0.7cm}
Let us define our \textit{database} $D$ as a collection of \textit{transactions} from the transactions set $T$: a transaction is defined as an element of our database, i.e., a sentence containing a textual description of the work dynamics leading to an accident. Hence, for each transaction we can define a set of words or a transformation of words, called \textit{items}, which belong to the items set $I$ and constitute our transaction. In this scenario, an association rule $A \Rightarrow B$ indicates that the occurrence of item $B$ is usually observed when also $A$ is found in the transaction, for $A, B \subset I$.

Association Rules mining procedure is based on three measures, namely \textit{Support}, \textit{Confidence}, and \textit{Lift}, whose values define whether a rule is meaningful or not. The support of an item $A$ can be defined as $Supp(A) = P(A) = \frac{\vert \{t | \forall t \in T \wedge A \in t\} \vert}{\vert T \vert}$
%
%
,namely, the percentage of transactions where item $A$ is found. Support can also be defined for a rule, e.g., $Supp(A \Rightarrow B)$, to represent the percentage of transactions in the database where both $A$ and $B$ are found. With the confidence of a rule we model the conditional probability of observing the consequent item having observed the antecedent, so it can hence be defined as $Conf(A \Rightarrow B) = \frac{P(AB)}{P(A)}$.
The third measure, i.e., the lift, is used to characterize the relationship between the antecedent and the consequent of the association rule. It is defined as
$Lift(A \Rightarrow B) = \frac{P(AB)}{P(A)P(B)} = \frac{P(B | A)}{P(B)}$, a lift greater than 1 identifies a positive relationship between the items, i.e., the conditional probability of observing $B$ given that $A$ is found is greater than the probability of observing $B$ in the database.

Apriori algorithm is a well-known association rules mining method in the literature, which leverages the measures we introduced to explore the itemset and build association rules from them~\cite{ref_apriori}. This method only focuses on the subset of those items belonging to the Frequent Itemset (FIS), i.e, those items whose support is greater than a minimum threshold $minsupp$. However, in some settings (e.g., medical descriptions) those items whose occurrence frequency is very low can be the ones carrying more semantical importance (e.g., names of diseases or even specific injuries). Following this intuition, we provided a Python implementation\footnote{The code is available at \url{https://github.com/AlessioMongelluzzo/FISinFIS_Apriori_Python}} to the algorithm FISinFIS Apriori proposed in~\cite{ref_fisinfis}, where Positive Association Rules (PARs) and Negative Association Rules (NARs) are built from both the Frequent Itemset and the Infrequent Itemset.
FISinFIS Apriori algorithm sets a minimum threshold on the Inverse Document Frequency (IDF), defined as $IDF(i) = \log \frac{\vert T \vert}{\vert \{t | t \in T \wedge i \in t\} \vert}$,
to filter those items which are too frequently used throughout the database. Moreover, we introduced in our implementation an additional threshold to the IDF as an upper bound for too rarely used items. The rationale behind this decision is to avoid typos and missing values placeholders which constitute a non-negligible part of our dataset.
The results coming from the application of the mining algorithm to our work accidents descriptions database allows RECKONition to detect those industrial properties or actions that are more likely related to an injury or, with negative rules, to their prevention, we show in Figure~\ref{fig:arresult} an AR graph example from a mock experiment performed on few sentences we wrote for this purpose.
%
\vspace{-0.2cm}
\begin{center}
    \includegraphics[width=0.75\textwidth]{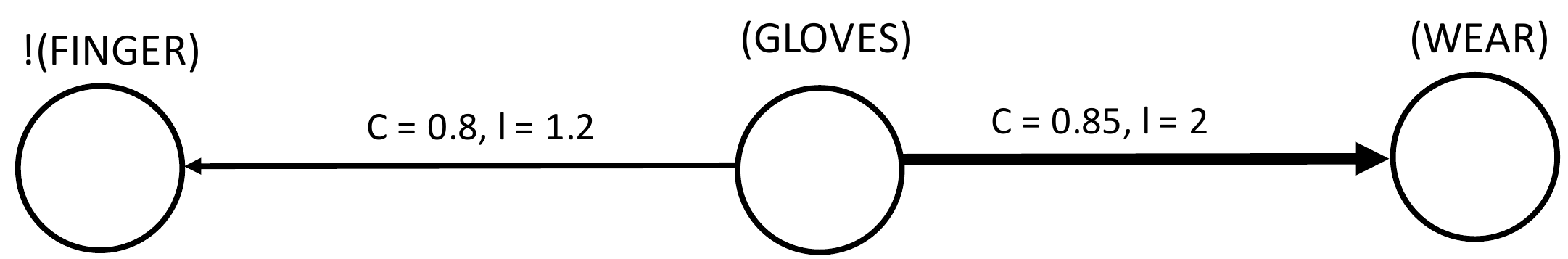}
    \vspace{-1cm}
    \begin{figure}[h]
    \caption{Example of an association rule graph from FISinFIS Apriori application.}
    \label{fig:arresult}
    \end{figure}
\end{center}
\vspace{-1.4cm}
\section{Natural Language Clustering }
\label{sct:naturallanguageclustering}
Textual Description Clustering~\cite{ref_dm} is performed to highlight differences and similarities among accident descriptions, and to group together the similar ones in order to facilitate the identification of appropriate interventions according to the situation described. In the following sections we will show the two different approaches used by our system. We will see both of them use the \textit{K-Medoids} algorithm~\cite{ref_kmedoid}, but differ in the operations performed on the \textit{dataset}.
%
\subsection{Tags Occurrence Clustering} 
\label{subsct:tagsoccurrenceclustering}
A first approach is based on an \textit{ontology} describing the context of interest (in the case of RECKONition, the metallurgical company) and uses TAGs representing ontological classes to integrate contextual knowledge in the clustering analysis. 

\begin{center}
    \includegraphics[width=0.7\textwidth]{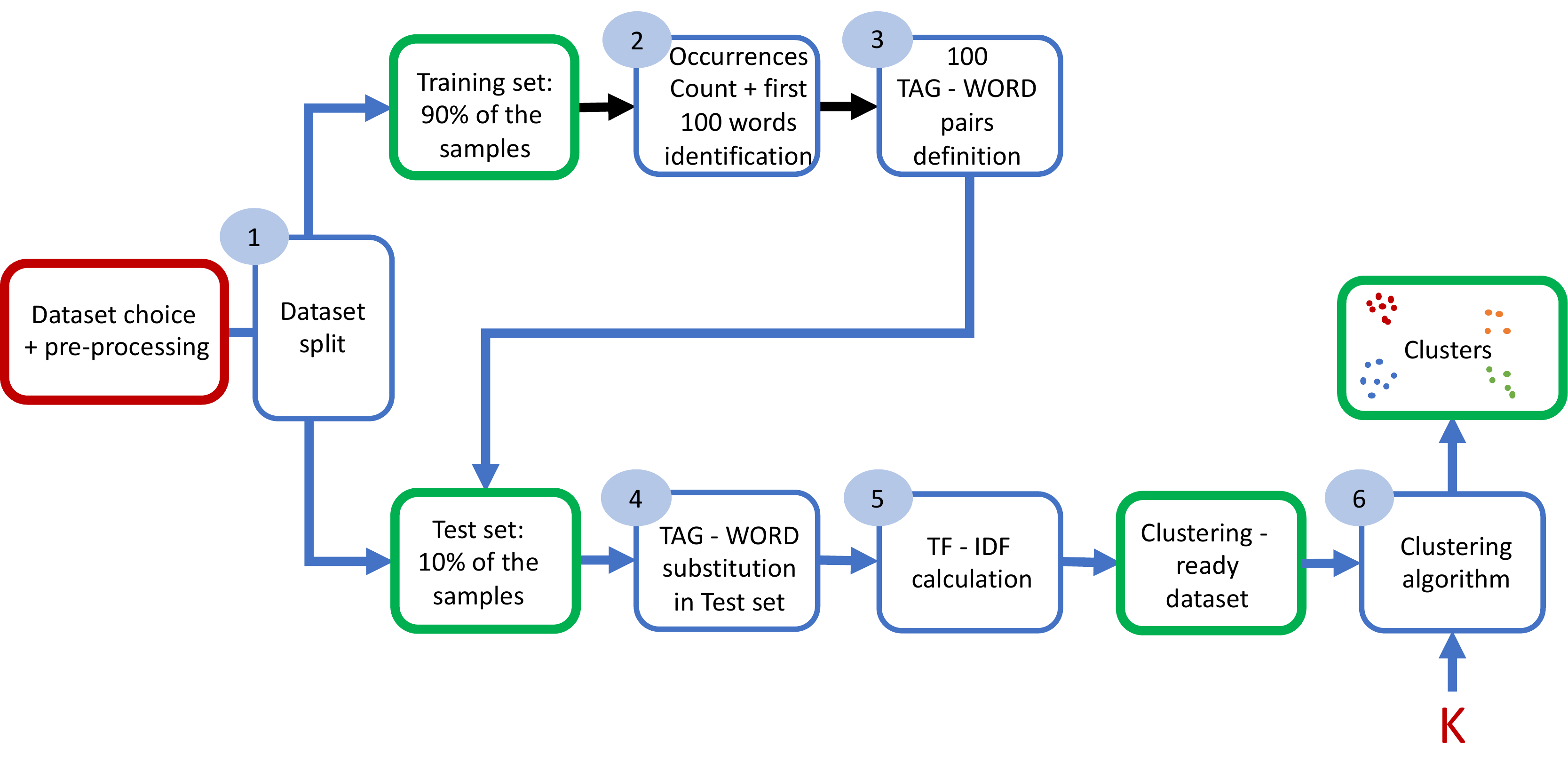}
    \vspace{-1cm}
    \begin{figure}[ht]
    \caption{Tags Occurrences Clustering - The process}
    \label{fig:process}
    \end{figure}
\end{center}
\vspace{-1cm}
As you can see in Figure~\ref{fig:process}, we calculate the occurrences of each word, we identify the hundred most frequent ones, and we define TAG - WORD pairs in the training set; we then perform the hundred substitution in the test set. In this way, words representing similar concepts are read by the algorithm as the same term. 
As can also be seen from the diagram, before calculating the occurrences, all the necessary pre-processing operations were carried on the entire \textit{dataset}, and in particular the stop\_words were eliminated. Almost all the remaining words resulted to be descriptive of the context then, and it was possible to associate them with a TAG. In the rare case in which this did not happen, or the substitution introduced ambiguity, the terms remained unchanged. Once performed the substitution, we calculate the Time frequency - Inverse Document Frequency (TF-IDF)~\cite{ref_dm} of each term in each description. After step 5 we than have a \textit{dataset} ready for clustering and it is composed by 6662 terms as features, whose values correspond to the calculated TF-IDF.

Once we have the \textit{dataset}, we perform \textit{K-Medoids}. Here we report some example of the clusters obtained by setting the $K$ $=$ \textit{n\_clusters} parameter equal to 30, and we show how we are able to separate the different situations described.
\vspace{-0.1 cm}
\begin{center}
    \includegraphics[width=0.9\textwidth]{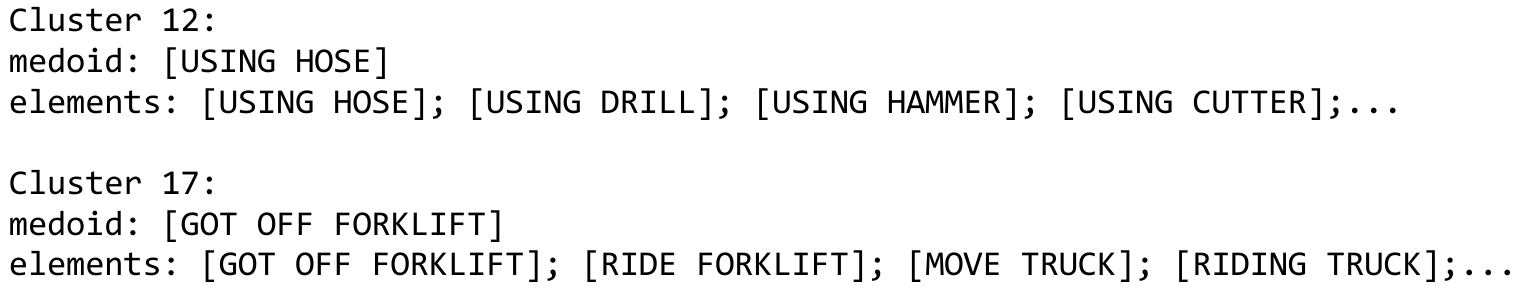}
    \vspace{-1 cm}
    \begin{figure}[h]
    \label{fig:occ_clustersample}
    \end{figure}
\end{center}
\vspace{-1 cm}

What is important to notice is that the algorithm is not only able to group together descriptions containing the exact same words, but it is also able to identify the ones concerning similar incidents in the analysed context. In this perspective, Clusters 12 represents accidents involving working tools, while Clusters 17 represents accidents occurring on board mobile machinery. 
\subsection{Transformers-Based Clustering} 
\label{subsct:transformersbasedclustering}
A different clustering approach adopted by RECKONition leverages the novel Encoder-Decoder models with \textit{Attention} and \textit{self-Attention} mechanism to build homogenous groups of accidents descriptions~\cite{ref_encdec}, \cite{ref_attention}. These techniques comprise several stacked encoder-decoder layers, which can model different syntactic structures with self-Attention modules together with feed-forward neural networks
, resulting in milions of trainable parameters: the base version of the Bidirectional Encoder Representations from Transformers (BERT) features 12 encoder-decoder layers, 768 hidden nodes for each layer and 12-attention-heads, resulting in 110M parameters, hence, requiring a very large amount of data to be properly trained, as well as time and computational capacity.
RECKONition uses a pre-trained BERT model publicly available\footnote{The pre-trained model can be found at \url{https://huggingface.co/dbmdz/bert-base-italian-cased}}, which was trained on 13GB of Italian textual data from Wikipedia and OPUS corpora\footnote{\url{https://opus.nlpl.eu/}}, and performs fine-tuning of such model on a subset of the accidents-at-work \textit{dataset} provided by INAIL. The fine-tuned model is then validated on a hold-out set which is used to perform clustering on unseen textual descriptions: during the forward step of a new sentence, the hidden states of the last decoder layer in the model are extracted and used as numerical features representing the sentence. This procedure is performed for each description in the validation set, so to build a new \textit{dataset} with shape $v \times (768*n_t)$, being 768 the number of hidden states of a BERT layer, $v$ the number of sentences in the validation set, and $n_t$ the number of length of the tokenized sentences. RECKONition leverages Incremental Principal Component Analysis (IPCA)~\cite{ref_ipca} to reduce the dimensionality of the representative features to a number of principal components so that at least 85\% of the variability of data is explained, and performs \textit{K-Medoids} clustering for values of the \textit{n\_clusters} $k$ in the range $[2, 100]$.
Unlike the clustering approach described in Section~\ref{subsct:tagsoccurrenceclustering}, this method can effectively find sentences with highly correlated syntactic structures. We show in the following two examples of clusters obtained with the Transformers-based clustering approach applied to a sample dataset.
\begin{center}
    \includegraphics[width=0.9\textwidth]{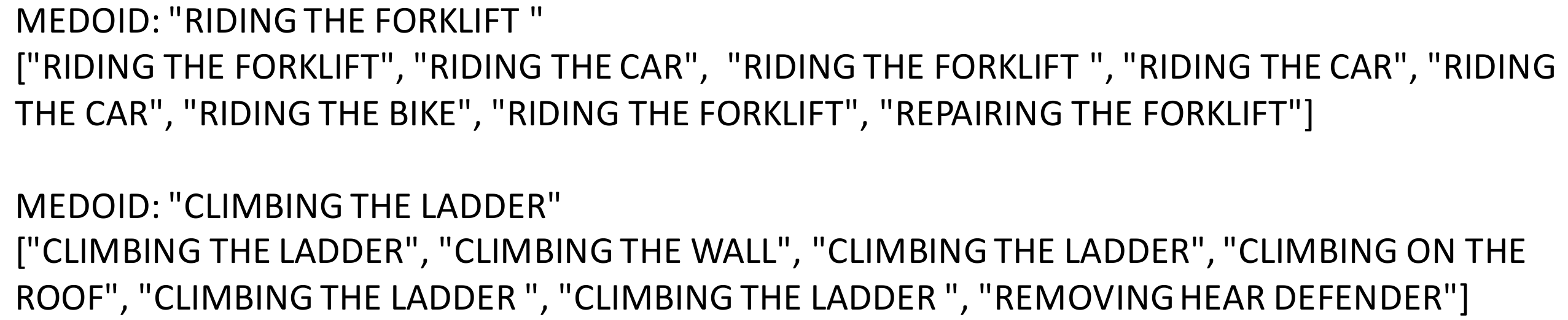}
    \vspace{-0.9cm}
    \begin{figure}[h]
    \label{fig:bertclustersample}
    \end{figure}
\end{center}
\vspace{-1.3cm}
%
%
%
\section{Natural Language Prediction}
\label{sct:naturallanguageprediction}
There is plenty of experimental settings where a Language Model (LM)~\cite{ref_languagemodel}, i.e., a model that is able to process textual sequences~\cite{ref_seqlearning}, can be effectively applied. These range from feature extraction (as we did with Trasformers-based clustering in Section~\ref{subsct:transformersbasedclustering}), to text classification, question answering, next sentence prediction, natural language inference, or even text generation. 
RECKONinition comprises a custom LM, which was properly built and trained on the accidents \textit{dataset} provided by INAIL, with the purpose of performing next sentence prediction from the textual description of the accidents dynamics to those of the consequences on the workers involved. The high-level architecture of the LM is shown in Figure~\ref{fig:lmschema}.
We emphasize that the tokenization step~\cite{ref_token}, which allows to move from symbolical (textual) to numerical representation, requires to define a vocabulary size to fix the number of tokens, which our model sets to 5000 items. The embedding layer is then used to project the sparse representation obtained by the tokenization step to a dense space with size 128. The sequence learning ability of the model resides in the \textit{n} layer shown in Figure~\ref{fig:lmschema}, which is composed of two Bidirectional Long Short-Term Memory (LSTM) with 100 units each~\cite{ref_lstm},~\cite{ref_bidirectional}, a flattening layer, two dense layers with 50 neurons each and ReLU activations, and a final Dropout layer with a 50\% drop rate before the output dense layer with 5000 neurons (one for each item) and sigmoidal activation. This model can be trained in a supervised fashion~\cite{ref_bishop} minimizing the Binary Cross Entropy loss function with Adaptive Moment Estimation (ADAM)~\cite{ref_adam}. This is achieved by providing the descriptions of the accidents dynamics as input features, together with the corresponding effects and injuries on the workers involved as input targets. For a new unseen sentence describing a working scenario, the model will output a probability vector on the items in the vocabulary defining those which better describe the consequence of the input description.
\begin{center}
    \includegraphics[width=0.8\textwidth]{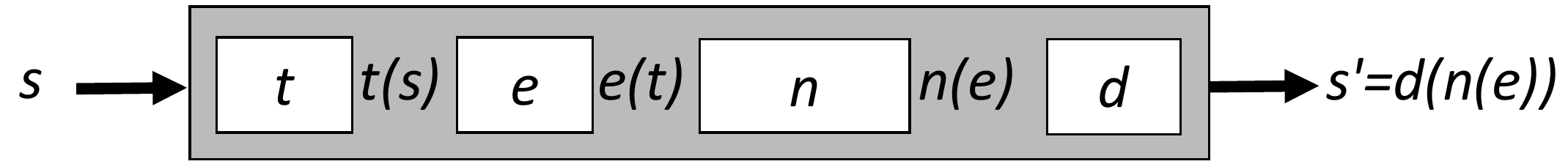}
    \vspace{-0.7cm}
    \begin{figure}[h]
    \caption{High-level architecture of the language model. \textit{s} and \textit{s'} are textual input and output, respectively. Layer \textit{t} represents the tokenizer, \textit{e} is an embedding layer, and \textit{d} a decoder. The main component \textit{n} is a recurrent neural network model.}
    \label{fig:lmschema}
    \end{figure}
\end{center}
\vspace{-0.8cm}
We show in in the following two examples of textual prediction given some input descriptions, where the output sentence results in a scattered sequence of items due to the limitation on the vocabulary size.
%
\begin{center}
    \includegraphics[width=0.7\textwidth]{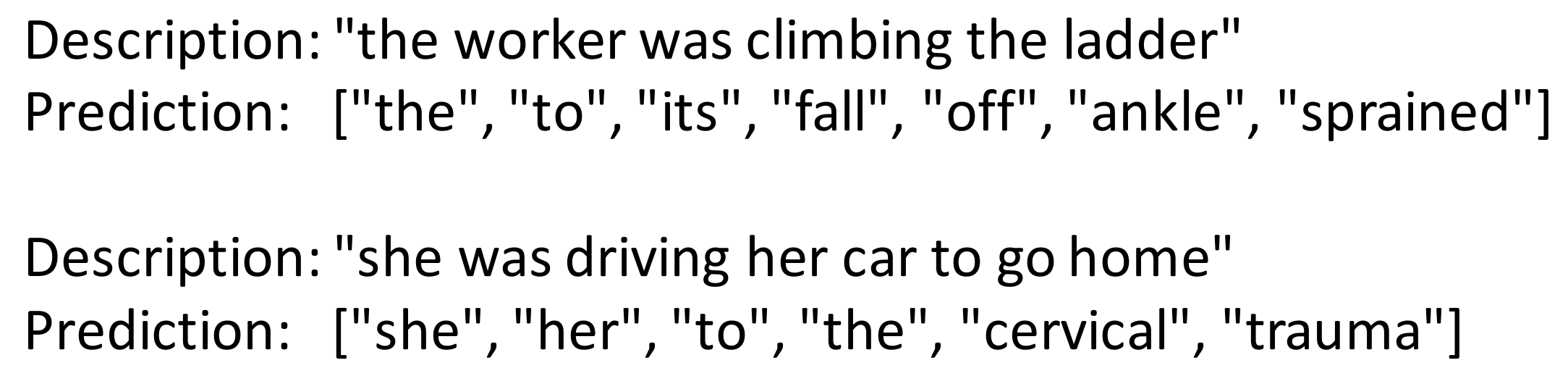}
    \vspace{-0.9cm}
    \begin{figure}[h]
    \label{fig:nlpredexamples}
    \end{figure}
\end{center}
\vspace{-1.3cm}
\section{Discussion and conclusion}
\label{sct:discussion_conclusion}
We introduced and described RECKONition system and its application scenario on a real world dataset. The system is currently operating to understand and extract knowledge from the accidents descriptions database and support the industries towards a safer workplace.
The results obtained and the continuous dialogue with INAIL's partners have highlighted the unsupervised ability of RECKONition to both cluster work activities and to identify relationship between accidents and consequences. As a next step, RECKONition will be deployed in real factories where proper sensors can produce inputs to the system, allowing real-time predictions to detect hazardous activities and at the same time improving data quality and performances of the system itself.

%
%
%
%

\end{document}